\documentclass{bmvc2k}

\usepackage{amsmath}
\usepackage{booktabs}
\usepackage{siunitx}
\usepackage{graphicx}  

\usepackage{subcaption}
\usepackage{colortbl}

\definecolor{gold}{RGB}{255, 215, 0}  
\definecolor{silver}{RGB}{192, 192, 192} 

\newcommand{\equalcontrib}{\textsuperscript{\dag}}

\title{PrIINeR: Towards Prior-Informed Implicit
Neural Representations for Accelerated MRI}

\addauthor{Ziad Al-Haj Hemidi\equalcontrib}{z.alhajhemidi@uni-luebeck.de}{1}
\addauthor{Eytan Kats\equalcontrib}{eytan.kats@uni-luebeck.de}{1}
\addauthor{Mattias P. Heinrich}{mattias.heinrich@uni-luebeck.de}{1}

\addinstitution{
 Institute of Medical Informatics \\
 University of Luebeck \\
 Ratzeburger Allee 160\\
 23562 Luebeck \\
 Germany
}

\runninghead{Al-Haj Hemidi, Kats, Heinrich
}{PrIINeR}

\begin{document}

\maketitle

\begingroup
\renewcommand\thefootnote{}
\footnotetext{\equalcontrib Equal contribution.}
\endgroup

\begin{abstract}
	Accelerating Magnetic Resonance Imaging (MRI) reduces scan time but often degrades image quality. While Implicit Neural Representations (INRs) show promise for MRI reconstruction, they struggle at high acceleration factors due to weak prior constraints, leading to structural loss and aliasing artefacts. To address this, we propose PrIINeR, an INR-based MRI reconstruction method that integrates prior knowledge from pre-trained deep learning models into the INR framework. By combining population-level knowledge with instance-based optimization and enforcing dual data consistency, PrIINeR aligns both with the acquired k-space data and the prior-informed reconstruction. Evaluated on the NYU fastMRI dataset, our method not only outperforms state-of-the-art INR-based approaches but also improves upon several learning-based state-of-the-art methods, significantly improving structural preservation and fidelity while effectively removing aliasing artefacts.
	PrIINeR bridges deep learning and INR-based techniques, offering a more reliable solution for high-quality, accelerated MRI reconstruction. The code is publicly available on \url{https://github.com/multimodallearning/PrIINeR}.
\end{abstract}

\section{Introduction}
\label{sec:intro}
Magnetic Resonance Imaging (MRI) is a widely used non-invasive imaging technique known for excellent soft tissue contrast, making it indispensable in medical diagnostics. However, conventional MRI scans often suffer from long acquisition times, leading to patient discomfort~\cite{griswold2002generalized,Lustig2007sparse,Pruessmann1999sense}. These limitations hinder the applicability of MRI for dynamic imaging and time-sensitive clinical scenarios. To address this challenge, accelerated imaging techniques have been extensively explored. One class of acceleration methods involves parallel imaging, which works by acquiring a reduced amount of k-space data with an array of receiver coils \cite{deshmane2012parallel}. While undersampling k-space allows for faster data acquisition, it introduces aliasing artifacts in the reconstructed images.

\textbf{Conventional parallel imaging methods}, including SENSE \cite{Pruessmann1999sense} and GRAPPA \cite{Griswold2002grappa}, address aliasing artifacts arising from undersampled k-space by solving an inverse problem constrained by coil sensitivity profiles. SENSE performs this reconstruction in the image domain, utilizing coil sensitivity maps to resolve aliasing. GRAPPA, in contrast, operates in k-space, estimating missing data points through kernel-based interpolation derived from autocalibration data.

\textbf{Compressed sensing-based MRI reconstruction}, building upon the principles of sparse signal recovery, offers substantial acceleration by exploiting prior knowledge of image sparsity \cite{Jung2007k,Lustig2007sparse}. These approaches formulate the reconstruction as an optimization problem, minimizing data inconsistency while enforcing sparsity through regularization terms, thereby enabling accurate image recovery from severely undersampled acquisitions.

\textbf{Data-driven techniques} have significantly advanced MRI reconstruction. Supervised learning methods, such as DeepCascade \cite{Schlemper2017deepcascade} and AUTOMAP \cite{Zhu2018automap}, train deep networks on extensive datasets to map undersampled k-space to fully sampled images, outperforming traditional methods in both speed and image quality. Hybrid approaches that integrate deep learning with physical models, such as MoDL \cite{Aggarwal2019modl} and Learned Primal-Dual \cite{Adler2018primaldual}, embed MRI physics constraints within deep learning architectures to enhance generalizability and robustness. Recent methods, including Reconformer \cite{guo2023reconformer} and GenINR \cite{li2023implicit}, leverage the latest advances in deep learning architectures to further push the boundaries of data-driven MRI reconstruction. While data-driven models trained on large populations are highly effective, they often prioritize generalization over preserving fine image details. This trade-off can lead to artifact-free but often smooth reconstructions, potentially sacrificing structures in individual images.

\textbf{Implicit Neural Representations} (INRs) have emerged as a promising paradigm for MRI reconstruction, parameterizing images as continuous functions represented by neural networks \cite{Mildenhall2020nerf,Sitzmann2020siren}. These methods have shown significant potential in reconstructing high-resolution images from sparsely sampled k-space data, enabling efficient instance-specific optimization without requiring extensive training datasets. Approaches like INR-based self-supervised learning \cite{Chen2021modinr,Tancik2020fourier} and motion-resolved INR models \cite{Zang2021motion} have expanded their applicability in medical imaging. While INRs excel at preserving fine structural details, they can struggle with complete artifact removal due to their inherent lack of population-level knowledge, leading to reconstructions that may retain residual artifacts.

\textbf{Contribution}. We introduce PrIINeR a \textbf{Pr}ior-\textbf{I}nformed \textbf{I}mplicit \textbf{Ne}ural \textbf{R}epresentation framework for MRI reconstruction that integrates population-level knowledge with the ability of INRs to preserve fine structural details. Our approach leverages a pre-trained deep learning model as a prior, which is then refined through instance-based optimization with a dual data consistency constraint. On one hand, data consistency enforces alignment with the prior knowledge provided by the deep learning model, while on the other, it incorporates information from the acquired undersampled k-space data. This enables effective knowledge transfer from the pretraining phase while maintaining adaptability and consistency at the instance level. The instance-based INR optimization ensures that the reconstruction remains faithful to the acquired k-space data, enhancing fine-detail preservation while benefiting from the artifact-free smoothness of population-pretrained models. Our method bridges the gap between deep learning-based and optimization-based reconstruction techniques by leveraging both population knowledge and instance-specific adaptation, providing a robust and generalizable solution for accelerated MRI. Our contributions can be summarized as follows:

\begin{itemize}
	\item[1.] We propose a novel MRI reconstruction framework that combines population-level knowledge from pre-trained deep learning models with instance-based optimization effectively. This approach enables reconstructions that are both smooth and artifact-free while preserving fine structural details.
	\item[2.] We introduce a dual data consistency objective in k-space, ensuring alignment with both the prior knowledge provided by the deep learning model and the acquired undersampled k-space data. This constraint enables effective knowledge transfer while maintaining fidelity to the acquired measurements.
	\item[3.] We demonstrate that the proposed method can be integrated in a plug-and-play manner with various population-based priors, not only enhancing fine-detail preservation in the reconstructed image but also boosting the performance of weak priors to state-of-the-art levels.
	\item[4.] We extensively evaluate our method and show, both quantitatively and qualitatively, that it outperforms state-of-the-art data-driven approaches as well as purely instance-based optimization methods. Our results highlight the advantages of leveraging both global priors and instance-specific refinements for improved MRI reconstruction quality.
\end{itemize}

\begin{figure}[h]
	\centering
	\includegraphics[width=\textwidth]{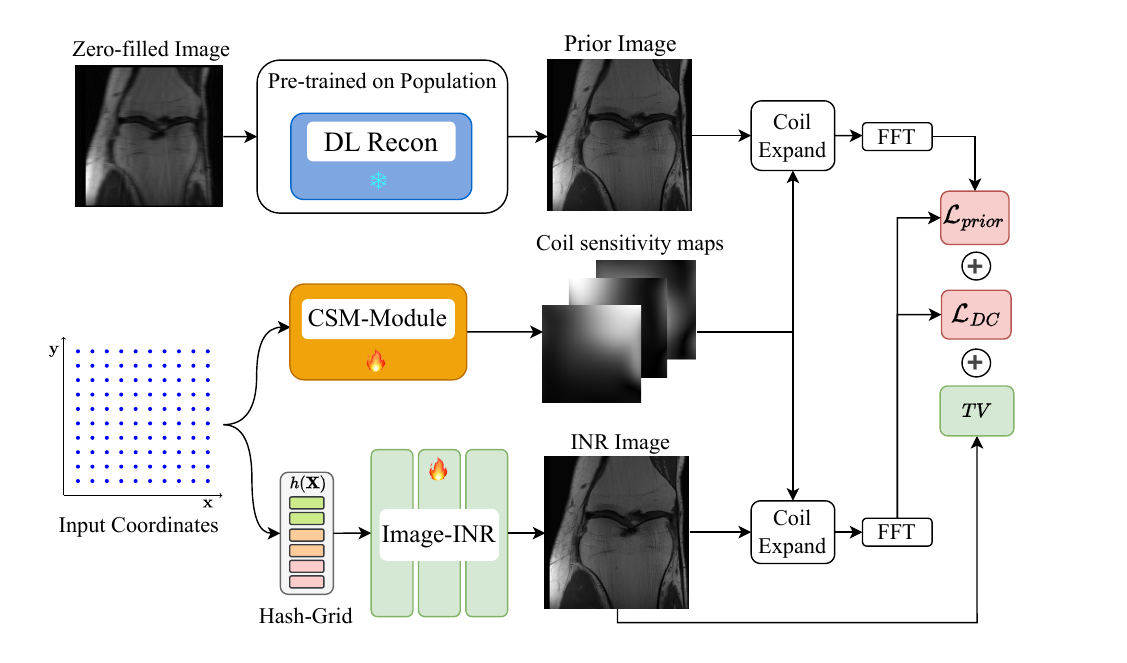}
	\caption{Overview of the proposed PrIINeR framework. A population-pretrained deep learning reconstruction model is used in a plug-and-play fashion to generate a prior image from the initial zero-filled input. Any state-of-the-art MRI reconstruction network can be substituted here. Instance-wise optimization then refines the Image-INR using hash-grid encoded coordinates by minimizing data consistency loss and prior k-space loss, with total variation regularization enforcing image-space sparsity. Coil sensitivity maps are estimated by optimizing the CSM-Module, which fits polynomial function coefficients.}
	\label{fig:overview}
\end{figure}

\section{Materials and Methods}
\label{sec:methods}
In this section, we present our proposed MRI reconstruction framework, which combines learning-based priors with INRs. We first introduce the problem formulation (Sec.~\ref{subsec:problem}), followed by a detailed explanation of our approach, including the dual data consistency constraint and the INR optimization (Sec.~\ref{subsec:method}). Further we describe the datasets used for evaluation and outline the experimental setup (Sec.~\ref{subsec:experiments}). An overview of the proposed method is illustrated in Fig.~\ref{fig:overview}.

\subsection{Problem Formulation}
\label{subsec:problem}

The MRI acquisition process can be described as a linear mapping from the image domain to k-space. Given an undersampled k-space measurements $\mathbf{y}$, the acquisition model is expressed as:
\begin{equation}
	\mathbf{y} = \textbf{M} \cdot \mathcal{F}(S \cdot \mathbf{x}) + \epsilon,
\end{equation}
where $\mathbf{x}$ represents the fully sampled image, $\textbf{M}$ is the undersampling mask, $\mathcal{F}$ is the Fourier transform operator, $S$ denotes the coil sensitivity maps, and $\epsilon$ accounts for complex-valued noise. The objective of MRI reconstruction is to estimate $\mathbf{x}$ from $\mathbf{y}$ by solving the inverse problem:
\begin{equation}
	\hat{\mathbf{x}} = \underset{\mathbf{x}}{\operatorname{argmin}} \; \underbrace{\left\| \mathbf{y} - \textbf{M} \cdot \mathcal{F}(S \cdot \mathbf{x}) \right\|_2^2}_{\text{acquired k-space consistency}} + \underbrace{\lambda \mathcal{R}(\mathbf{x})}_{\text{regularization}},
\end{equation}
where $\mathcal{R}(\mathbf{x})$ enforces prior knowledge, and $\lambda$ controls the trade-off between data fidelity and regularization. Common choices for $\mathcal{R}$ include sparsity-based constraints, total variation, or deep learning priors. In some approaches, the sensitivity maps are estimated jointly with the image \cite{tang2023jsense}, whereas others assume them to be precomputed \cite{Pruessmann1999sense}.

\subsection{Proposed Method}
\label{subsec:method}

We propose \textbf{PrIINeR}, a \textbf{Pr}ior-\textbf{I}nformed \textbf{I}mplicit \textbf{Ne}ural \textbf{R}epresentation framework for MRI reconstruction, which synergistically combines deep learning-based priors with INRs. Our approach simultaneously optimizes a multi-layer perceptron (MLP) to generate the reconstructed MRI image and polynomial function coefficients representing the coil sensitivity maps. A dual data consistency constraint ensures alignment with both the acquired k-space data and prior information.

\subsubsection{Joint Optimization of Coil Sensitivity Maps and Image Reconstruction.}
\label{subsubsec:joint_optimization}

In the instance optimization process, we represent the MRI image intensity using an implicit neural representation (INR) with hash-grid encoding~\cite{muller2022instant} of spatial coordinates. The INR is implemented as a multi-layer perceptron (MLP) parameterized by learnable weights $\theta$ \cite{feng2023imjense}. Simultaneously, we optimize the polynomial coefficients $\varphi$ that define the coil sensitivity maps \cite{ying2007joint}. This joint optimization approach ensures that the reconstructed image remains consistent with both the acquired k-space measurements and the estimated sensitivity maps.

The acquired k-space consistency is enforced through the following loss function:

\begin{equation}
	\mathcal{L}_{\text{DC}}(I(\theta), \varphi) = \sum _{j = 1}^{c} \left\| \mathbf{y}_{j} - \textbf{M} \cdot \mathcal{F} \big( S_{j}(\varphi) \cdot I(\theta) \big) \right\|_2^2.
\end{equation}

Here, $\mathbf{y}_{j}$ represents the acquired k-space data for coil $j$, $I(\theta)$ denotes the INR-based image reconstruction parameterized by $\theta$, and $S_{j}(\varphi)$ corresponds to the estimated coil sensitivity map parameterized by $\varphi$. Sensitivity maps weight different spatial regions, disentangling coil-specific image contributions.

\subsubsection{Dual Data Consistency.}
\label{subsubsec:data_consistency}

To effectively incorporate population-based knowledge, we introduce an additional data consistency term that enforces alignment between the estimated reconstruction $I(\theta)$ and the prior-based reconstruction $\mathbf{\hat{x}_p}$ obtained from a pre-trained model. This consistency is enforced in k-space as follows:

\begin{equation}
	\mathcal{L}_{\text{prior}}(I(\theta),\varphi) = \sum _{j = 1}^{c} \left\| \mathcal{F} \big( S_{j}(\varphi) \cdot \mathbf{\hat{x}_p} \big) - \mathcal{F} \big( S_{j}(\varphi) \cdot I(\theta) \big) \right\|_2^2.
\end{equation}

The overall optimization objective is then formulated as:

\begin{equation}
	\label{eq:total_loss}
	\mathcal{L}(I(\theta), \varphi) = \underbrace{\alpha \mathcal{L}_{\text{DC}}(I(\theta), \varphi)
	+ \mathcal{L}_{\text{prior}}(I(\theta),\varphi)}_{\text{k-space}}
	+ \underbrace{\lambda \mathcal{R} (I(\theta))}_{\text{image space}}.
\end{equation}

Here, $\alpha$ is a weighting factor that balances the trade-off between acquired k-space consistency and prior-based consistency, while $\lambda$ controls the strength of the regularization. This formulation ensures that the reconstructed image $I(\theta)$ remains consistent with both the acquired measurements and the learned population prior while incorporating regularization to suppress aliasing artifacts caused by undersampling.

As the regularization term $\mathcal{R}(I(\theta))$, we employ the Total Variation (TV) objective, which promotes piecewise smoothness by penalizing abrupt intensity variations while preserving important structural details and edges in the reconstructed image.

\subsection{Experiments}
\label{subsec:experiments}

\textbf{Datasets.} We evaluate our method on the NYU fastMRI dataset\footnote{\url{fastmri.med.nyu.edu}}~\cite{knoll2020fastmri,zbontar2018fastmri}, using the 3D 15-coil knee k-space dataset with fully sampled k-space. The dataset is split into 200/50/50 volumes for training, validation, and testing. Images are center-cropped to $\text{\#slices} \times 320 \times 320$ before being transformed back to k-space. We retrospectively apply equispaced 2D Cartesian undersampling on-the-fly with acceleration factors of 4, 6, 8, and 10, retaining 8\% of the k-space center.

\noindent\textbf{Experimental setup.} We evaluate our method against state-of-the-art MRI reconstruction approaches, including INR- and deep learning-based methods. The benchmarked methods include Zero-filled, IMJENSE Acoeff~\cite{feng2023imjense}, IMJENSE Hash~\cite{feng2023imjense}, U-Net~\cite{myronenko20193d}, ReconFormer~\cite{guo2023reconformer}, and GenINR~\cite{li2023implicit}.

IMJENSE Acoeff and IMJENSE Hash are instance-based INR methods estimating coil sensitivity maps using polynomial fitting or hash-grid encoding MLP, with $\lambda$ set to $1e^{-4}$. U-Net, ReconFormer, and GenINR are population-trained baseline methods. GenINR integrates an INR with a convolutional neural network CNN-based feature extractor. Our implementation of IMJENSE is based on \cite{feng2023imjense}.
GenINR and ReconFormer were trained using the configurations provided in their official repository \footnote{\url{https://github.com/YuSheng-Zhou/Generalized_INR.git}}. For U-Net, we implemented a 2D \emph{SegResNet} from MONAI\footnote{\url{https://monai.io/}}, setting the \texttt{spatial\_dims} to 2 and \texttt{in\_channels} to 1. It was trained using the Adam optimizer for 500 epochs with a learning rate of $1e^{-3}$ and a batch size of 25 until convergence. All deep learning models were trained on 2D slices extracted from the 3D dataset.

PrIINeR is instance-based and can be combined with any population-based deep learning prior. To demonstrate its flexibility, we experiment with different pre-trained models from the comparison methods as priors for PrIINeR, namely PrIINeR-UNet, PrIINeR-GenINR, and PrIINeR-ReconFormer. We use Adam for optimization (learning rate = $1e^{-2}, \quad \lambda = 10^{-4}, \quad \alpha = 0.8$). We use a 3 layer ReLU MLP with 64 hidden units and the default hash-grid encoding settings~\cite{muller2022instant}. The input is a 2D coordinate grid, while the output of the MLP is a real-valued intensity image.

All methods were implemented in PyTorch\footnote{\url{https://pytorch.org/}} and executed on a single NVIDIA A100-80GB GPU. We report the Structural Similarity Index (SSIM) and Peak Signal-to-Noise Ratio (PSNR) as evaluation metrics. The code is publicly available on \url{https://github.com/multimodallearning/PrIINeR}.

\begin{figure}[h!]
	\centering
	\includegraphics[width=\textwidth]{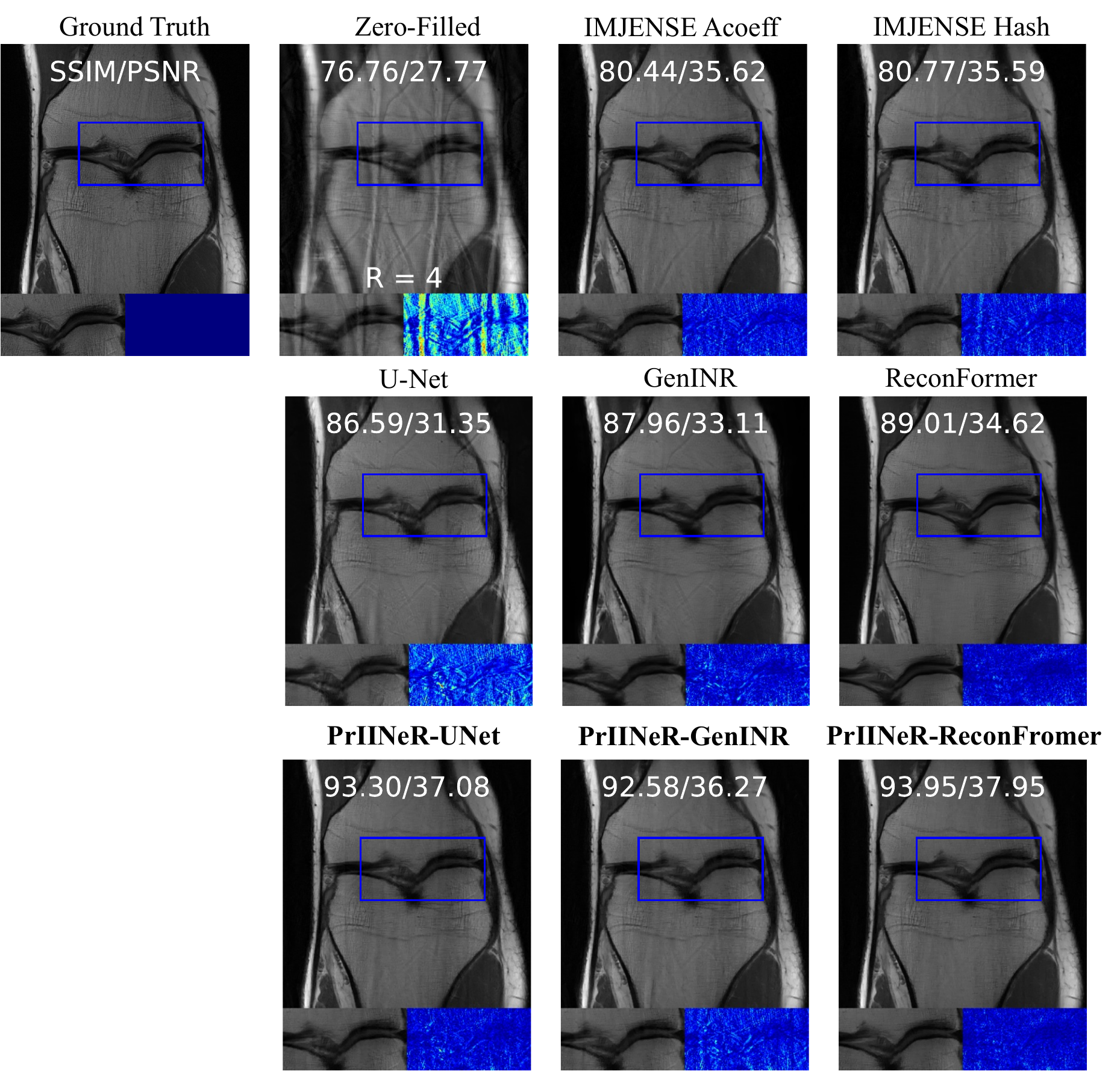}
	\vspace*{6pt}
	\caption{Comparison of the state-of-the-art to our method PrIINeR at acceleration factor 4. We report the SSIM (\%) and PSNR (db) values at the top of each image for each method. The lower part of each image presents a zoomed-in view of the blue-boxed region from the upper part, along with the corresponding difference map to the ground truth crop (Jet colormap in which min/max values are in range of [0, 1]).}

	\label{fig:PrIINeRAcc04}
\end{figure}

\begin{figure}[h!]
	\centering
	\includegraphics[width=\textwidth]{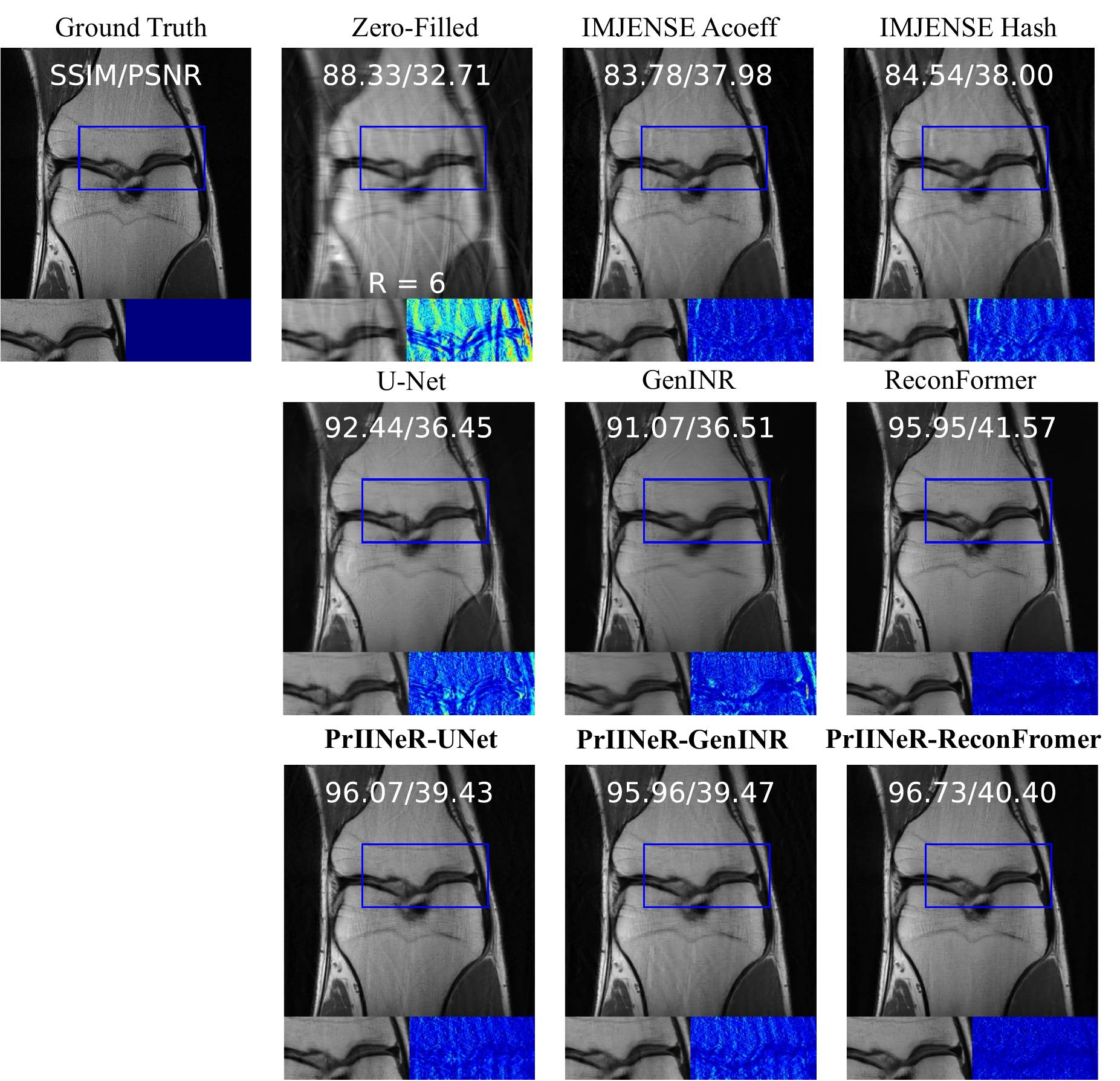}
	\vspace*{6pt}
	\caption{Comparison of the state-of-the-art to our method PrIINeR at acceleration factor 6. We report the SSIM (\%) and PSNR (db) values at the top of each image for each method. The lower part of each image presents a zoomed-in view of the blue-boxed region from the upper part, along with the corresponding difference map to the ground truth crop (Jet colormap in which min/max values are in range of [0, 1]).}

	\label{fig:PrIINeRAcc06}
\end{figure}

\begin{table}[t]
	\centering
	\caption{Quantitative results on the fastMRI Knee test set across acceleration rates. Metrics are reported as SSIM (\%) and PSNR (dB), shown as \(\mu_{\pm \sigma}\). \colorbox{gold}{Gold} marks indicate where PrIINeR outperforms the DL prior model, while \colorbox{silver}{silver} marks indicate lower performance.} 
	\vspace*{6pt} 
	\label{tab:quant_results}
	\renewcommand{\arraystretch}{1.2} 
	\vspace*{2pt} 
	\resizebox{\textwidth}{!}{ 
		\begin{tabular}{lcccccccc}
			\toprule
			                                      & \multicolumn{4}{c}{\textbf{SSIM}$\uparrow$ (\(\%\))} & \multicolumn{4}{c}{\textbf{PSNR}$\uparrow$ (dB)}                                                                                                                                                                     \\
			\cmidrule(lr){2-5} \cmidrule(lr){6-9}
			Method                                & R = 4                                                & R = 6                                            & R = 8                                    & R = 10                                                 & R = 4                                & R = 6 & R = 8 & R = 10 \\

			\midrule
			Zero-filled                           & 80.02\(_{\pm4.22}\)                                  & 76.19\(_{\pm5.39}\)                              & 72.70\(_{\pm6.18}\)                      & 70.28\(_{\pm6.78}\)                                    &
			31.26\(_{\pm1.66}\)                   & 30.63\(_{\pm1.67}\)                                  & 30.08\(_{\pm1.68}\)                              & 29.67\(_{\pm1.68}\)                                                                                                                                               \\
			\hline
   
			IMJENSE Acoeff~\cite{feng2023imjense} & 85.95\(_{\pm5.47}\)                                  & 84.27\(_{\pm5.78}\)                              & 80.93\(_{\pm6.36}\)                      & 79.06\(_{\pm6.32}\)                                    &
			36.90\(_{\pm3.16}\)                   & 35.66\(_{\pm2.89}\)                                  & 34.03\(_{\pm2.15}\)                              & 32.94\(_{\pm1.96}\)                                                                                                                                               \\
			IMJENSE Hash~\cite{feng2023imjense}   & 86.38\(_{\pm5.22}\)                                  & 84.88\(_{\pm5.44}\)                              & 81.59\(_{\pm5.71}\)                      & 79.64\(_{\pm5.56}\)                                    &
			36.83\(_{\pm2.93}\)                   & 35.69\(_{\pm2.56}\)                                  & 33.91\(_{\pm1.94}\)                              & 32.67\(_{\pm1.73}\)                                                                                                                                               \\
			\hline
			U-Net~\cite{griswold2002generalized}  & 88.79\(_{\pm2.74}\)                                  & 89.67\(_{\pm2.61}\)                              & 86.37\(_{\pm3.15}\)                      & 85.71\(_{\pm3.29}\)                                    &
			33.34\(_{\pm1.80}\)                   & 33.97\(_{\pm1.87}\)                                  & 32.27\(_{\pm1.81}\)                              & 31.93\(_{\pm1.86}\)                                                                                                                                               \\
			GenINR~\cite{li2023implicit}                                & 90.82\(_{\pm3.03}\)                                  & 90.27\(_{\pm3.30}\)                              & 87.99\(_{\pm3.80}\)                      & 87.01\(_{\pm3.95}\)                                    &
			35.54\(_{\pm1.73}\)                   & 35.21\(_{\pm1.77}\)                                  & 33.75\(_{\pm1.73}\)                              & 33.04\(_{\pm1.74}\)                                                                                                                                               \\
			ReconFormer~\cite{guo2023reconformer} & 93.66\(_{\pm2.20}\)                                  & 94.99\(_{\pm1.93}\)                              & 91.30\(_{\pm2.65}\)                      & 90.44\(_{\pm2.81}\)                                    &
			38.40\(_{\pm3.47}\)                   & 39.89\(_{\pm1.83}\)                                  & 36.51\(_{\pm1.74}\)                              & 35.86\(_{\pm1.74}\)                                                                                                                                               \\
			\hline

			\textbf{PrIINeR-UNet}                 & \cellcolor{gold} 93.38\(_{\pm2.74}\)                 & \cellcolor{gold}   94.70\(_{\pm2.19}\)           & \cellcolor{gold}     89.67\(_{\pm3.19}\) & \cellcolor{gold}                   88.55\(_{\pm3.42}\) & \cellcolor{gold} 38.28\(_{\pm2.03}\)
			                                      & \cellcolor{gold} 39.44\(_{\pm2.05}\)                                  & \cellcolor{gold}    35.29\(_{\pm1.74}\)          & \cellcolor{gold}    34.34\(_{\pm1.69}\)                                                                                                                           \\
			\textbf{PrIINeR-GenINR}               & \cellcolor{gold} 93.29\(_{\pm2.48}\)                 & \cellcolor{gold}   94.30\(_{\pm2.20}\)           & \cellcolor{gold}   89.26\(_{\pm3.19}\)   & \cellcolor{gold}        87.85\(_{\pm3.45}\)            & \cellcolor{gold} 37.87\(_{\pm1.94}\)
			                                      & \cellcolor{gold}  38.85\(_{\pm1.97}\)                & \cellcolor{gold} 34.76\(_{\pm1.77}\)             & \cellcolor{gold}   33.64\(_{\pm1.74}\)                                                                                                                            \\

			\textbf{PrIINeR-ReconFormer}          & \cellcolor{gold} 94.32\(_{\pm2.41}\)                 & \cellcolor{gold} 95.23\(_{\pm2.19}\)             & \cellcolor{silver} 91.21\(_{\pm2.95}\)   & \cellcolor{silver} 90.20\(_{\pm3.11}\)                 &
			\cellcolor{gold} 39.35\(_{\pm2.13}\)  & \cellcolor{gold} 40.42\(_{\pm2.26}\)                 & \cellcolor{gold} 36.53\(_{\pm1.87}\)             & \cellcolor{silver} 35.60\(_{\pm1.80}\)                                                                                                                            \\

			\bottomrule
		\end{tabular}
	}
\end{table}

\section{Results \& Discussion}
\label{sec:results}

Table~\ref{tab:quant_results} presents the quantitative comparison with state-of-the-art methods. Our method consistently improves SSIM and PSNR across all acceleration rates, with the exception of ReconFormer, which slightly outperforms our approach at the highest acceleration rate. Fig.~\ref{fig:PrIINeRAcc04} and Fig.~\ref{fig:PrIINeRAcc06} show visual results for acceleration factor 4 and 6, where all methods improve over the zero-filled baseline. Among them, ReconFormer and our hybrid variant, PrIINeR-ReconFormer, yield the most visually convincing reconstructions.

PrIINeR effectively preserves fine structural details while reducing aliasing artefacts more consistently than IMJENSE, U-Net, and GenINR. Notably, PrIINeR elevates the performance of weak deep priors like the U-Net to levels that closely rival state-of-the-art methods, achieving results comparable to the plain ReconFormer. Our improvements are statistically significant (p < 0.05) in 11 out of 12 SSIM comparisons and 10 out of 12 PSNR comparisons across 3 methods and 4 acceleration rates, using the Wilcoxon signed-rank test. These results demonstrate the robustness of our approach under diverse conditions.

A key strength of PrIINeR is its flexible design, which intentionally bridges deep learning priors with instance-based INR optimization. Population-trained priors offer global anatomical and structural guidance, while the instance optimization of the INR ensures precise adaptation to each specific scan without requiring supervised labels. This combination brings out the best of both paradigms, leveraging the generalization and speed of deep learning while retaining the fidelity and flexibility of INRs. By design, PrIINeR alleviates limitations from both sides, avoiding hallucination and over-smoothing typical of deep networks, and correcting the structure-inconsistency and under-constrained behavior of standalone INRs.

The trade-off between data fidelity and prior consistency is controlled by the weighting parameter $\alpha$ in Eq.~\ref{eq:total_loss}. While a full sensitivity analysis is beyond the current scope, we empirically chose an $\alpha$ that balances artefact suppression with fine detail retention. Excessively high values risk over-regularizing the reconstruction, while low values fail to suppress aliasing effectively. In future work, we plan to conduct a thorough ablation study over $\alpha$ and other hyperparameters to optimize performance across varied anatomical regions and sampling strategies.

We focused our comparisons on recent and relevant baselines, including ReconFormer, a modern unrolled transformer-based architecture, and GenINR, a generalized implicit neural representation, as both represent the current state-of-the-art. This choice reflects our goal of positioning PrIINeR in the context of cutting-edge reconstruction techniques that balance expressiveness, efficiency, and generalization.

Looking forward, we plan to extend PrIINeR by integrating more generalizable pre-trained priors that work across anatomies, contrasts, and k-space trajectories. The methods modularity and flexibility also make it suitable for incorporating additional priors such as physical models, motion dynamics, or contrastive representations, paving the way toward a more interpretable and clinically robust reconstruction pipeline for accelerated MRI.
It is worth noting that PrIINeR demonstrates promising potential for hybrid optimization beyond MRI reconstruction, with applicability to a broad spectrum of inverse problems in medical and computational imaging.

\section{Conclusion}
\label{sec:conclusion}
We presented PrIINeR, a Prior-Informed Implicit Neural Representations framework for MRI reconstruction that integrates population-based priors with INRs in a plug-and-play manner. Our method significantly improves structural preservation and fidelity where traditional INR approaches struggle. By leveraging population-based priors in a dual data-consistency, PrIINeR offers a more reliable and generalizable solution for high-quality, accelerated MRI reconstruction, effectively bridging the gap between deep learning-based and optimization-based techniques.

\bibliography{references}

\end{document}